\documentclass[runningheads,letter]{llncs}

\usepackage[utf8]{inputenc} 
\usepackage[T1]{fontenc}    
\usepackage{hyperref}       
\usepackage{url}            
\usepackage{booktabs}       
\usepackage{amsfonts}       
\usepackage{nicefrac}       
\usepackage{microtype}      
\usepackage{amsfonts}       
\usepackage{nicefrac}       
\usepackage{microtype}      
\usepackage{pifont,xspace,epsfig,wrapfig,paralist}
\usepackage{amsmath,amssymb,times}
\usepackage{latexsym,paralist,multirow,tablefootnote}
\usepackage{hyperref}
\hypersetup{colorlinks=true,linkcolor=blue,citecolor=blue}
\usepackage{graphicx}
\usepackage{amsfonts}
\usepackage{makecell}
\usepackage{graphicx}
\usepackage{adjustbox}
\usepackage{subcaption}
\usepackage{wrapfig,lipsum,booktabs}
\usepackage[]{algorithm2e}
\usepackage{graphicx}
\usepackage{wrapfig}
\usepackage{caption}

\newcommand{\BNN}{\mbox{\sc Bnn}}
\newcommand{\BBNN}{\mbox{\sc EncBNN}}

\newcommand{\Map}{\mbox{\sc M}}
\newcommand{\neub}{\mbox{\sc Nb}}
\newcommand{\prt}{\mbox{\sc Pr}}

\begin{document}
\mainmatter  

\title{Constrained Image Generation Using Binarized Neural Networks with Decision Procedures}
\titlerunning{Constrained Image Generation}
\author{
  {Svyatoslav Korneev}\inst{1}
  \and
  {Nina Narodytska}\inst{2}
  \and
  {Luca Pulina}\inst{3}
  \and
  {Armando  Tacchella}\inst{4}
  \and
  {Nikolaj  Bjorner}\inst{5}
  \and
  {Mooly    Sagiv}\inst{2}\inst{6}
}

\authorrunning{Svyatoslav Korneev \emph{et al.}}
\institute{Department of Energy Resources Engineering, Stanford University, Stanford, CA
  94305, USA
  \and
  VMware Research, Palo Alto, CA, USA
  \and
  Chemistry and Pharmacy Dept., University of Sassari, Via Vienna 2,
  Sassari, Italy
  \and
  DIBRIS, University of Genoa, Viale Causa 13, 16145
  Genoa, Italy
  \and
  Microsoft Research, One Microsoft Way, Washington, USA
  \and
  Tel Aviv University
}

\maketitle
\vspace{-10pt}
\begin{abstract}
  We consider the problem of  binary image generation with given properties. This problem arises in a number of practical applications, including generation of artificial porous medium for an electrode of lithium-ion batteries, for composed materials, etc. A generated image represents a porous medium and, as such, it is subject to two sets of constraints: topological constraints on the structure and process constraints on the physical process over this structure.
To perform image generation we need to define a mapping from a porous medium to its physical process parameters.
For a given geometry of a porous medium, this mapping can be done by solving a partial differential equation (PDE).
However, embedding a PDE solver into the search procedure is computationally expensive. We use a binarized neural network to approximate a PDE solver. This allows us to encode the entire problem as a logical formula.
Our main contribution is that, for the first time,  we show that this problem can be tackled using decision procedures. Our experiments show that our model is able to produce random constrained images that satisfy both topological and process constraints.
%

\end{abstract}
\section{Introduction}
We consider the problem of constrained image generation of a porous medium with given properties.
Porus media occur, e.g., in lithium-ion batteries and composed
materials~\cite{7962936,Ilenia11}; the problem of generating porus media with a given
set of properties is relevant in practical applications of material
design~\cite{Hermann,Pyrcz,Hornung:1996}.
Artificial porous media are useful during the manufacturing process as they allow the designer to
synthesize new materials with predefined properties. For example, generated images can be used in designing a new porous medium for an electrode of lithium-ion batteries. It is well-known that ions macro-scale transport and reactions rates are sensitive to the topological properties of the porous medium of the electrode. Therefore, manufacturing the porous electrode with given properties allows  improving the battery performance~\cite{7962936}.

\begin{figure}
\centering
\includegraphics[width=1\linewidth]{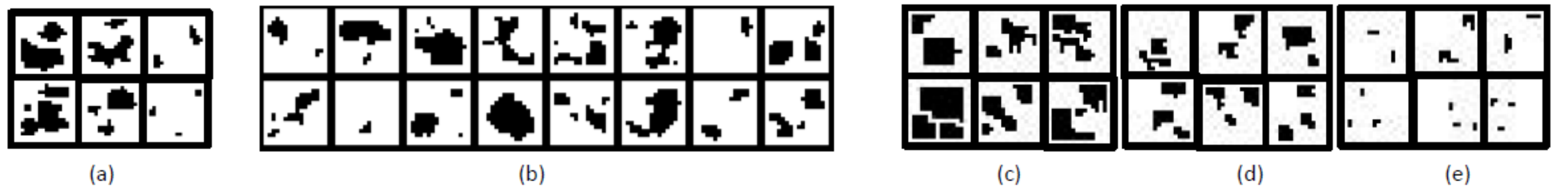}
\vspace{-7pt}
\caption{{\small (a) Examples of images from train sets with two and three grains; (b) Examples of images generated by a GAN on the dataset with two grains. Examples of generated images with (c) $d \in [40,50)$,  (d) $d \in [60,70)$, and  (e) $d \in [90,100]$.}}
\label{fig:images}
\end{figure}

Images of porous media\footnote{Specifically, we are looking at a transitionally  periodic ``unit cell'' of porous medium assuming that porous medium has a periodic structure~\cite{Hornung:1996}.} are black and white images that represent an abstraction of the physical structure.
Solid parts (or so called grains) are encoded as a set of connected black pixels; a void area is encoded a set of connected white pixels.
There are two important groups of restrictions  that images of a porous medium have to satisfy.
The first group constitutes a set of  ``geometric'' constraints that come from the problem domain and
control the total surface area of grains.
For example, an image contains two isolated solid parts.
Figure~\ref{fig:images}(a) shows examples of 16x16 images from our datasets with two (the top row) and three (the bottom row) grains.
\vspace{-10pt}
\vspace{-20pt}
The second set  of restrictions comes from the physical process that is defined for the corresponding porous medium.
In this paper, we consider  the macro-scale transportation process that can be described by a set of dispersion coefficients depending on the transportation direction. For example, we might want to generate images that have two grains such that  the dispersion coefficient along the $x$-axis is between 0.5 and 0.6.  The dispersion coefficient is defined for the given geometry of a porous medium. It can be obtained as a numerical solution of the diffusion Partial Differential Equation (PDE). We refer to these restrictions on the parameters of the physical process as process constraints.

The state of the art approach to generating synthetic images is to use generative adversarial networks (GANs)~\cite{GoodfellowPMXWOCB14}. However,
GANs are not able learn geometric, three-dimensional perspective, and counting constraints which is a known issue with this approach~\cite{Goodfellow17,Osokin}. Our experiments with GAN-generated images also reveal this problem. There are no methods that allow
embedding of declarative constraints in the image generation  procedure at the moment.

In this work we show that the image generation problem can be solved using decision procedures for  porous media.
We show that both geometric and process constraints can be encoded as a logical formula. Geometric constraints are encoded
as a set of linear constraints. To encode process constraints, we first approximate  the diffusion PDE solver with a Neural Network(NN)~\cite{Korneev1,Korneev2}. We use a special class of NN, called $\BNN$, as these networks can be  encoded as logical formulas. Process constraints are encoded as restrictions on outputs of the network.
This provides us with an encoding of the image generation problem as a single logical formula.
The contributions of this paper can be summarized as follows:
(i)~We show that constrained image generation can be encoded as a logical formula and tackled using decision procedures. (ii)~We experimentally investigate a GAN-based approach to constrained image generation
and analyse their advantages and disadvantages compared to the constraint-based approach.
(iii)~We demonstrate that our constraint-based approach is capable of generating random images that have given properties, i.e., satisfy  process constraints.

\vspace{-10pt}
\section{Problem description}
\vspace{-10pt}

We describe a constrained image generation problem.
We denote $I \in \{0,1\}^{t \times t}$ an image that encodes a porous medium and $d \in \mathbb{Z}^m$ a vector
of parameters of the physical process defined for this porous material.
We use an image and a porous medium interchangeably to refer to $I$.
We assume that there is a mapping function $\Map$ that maps an image $I$ to the corresponding parameters
vector $d$, $\Map: I \rightarrow   \mathbb{Z}^m$. We denote as $C_g(I)$ the geometric constraints on the
structure of the image $I$ and as $C_p(d)$ the process constraints on the vector of parameters $d$.
Given  a set of geometric and process constraints and a mapping function
$\Map$, we need to generate a random image $I$ that satisfies $C_g$ and $C_p$.
Next we overview geometric and process constraints and discuss the mapping function.

The geometric constraints $C_g$ define a topological structure of the image. For example, they can ensure that a given number of grains is present on an image and these grains do not overlap. Another type of constraints focuses on a single grain. They can restrict the shape of a grain, e.g., a convex grain, its size or position on the image. The third type of constraints are boundary constraints that ensure that the boundary of the image must be in a void area.  Process constraints define restrictions on the vector of parameters. For example, we might want to generate images with $d_i^j \in [a_j,b_j]$, $j = 1,\ldots,m$.

Next we consider a mapping function $\Map$. A standard way to define $\Map$ is by solving a system of partial differential equations. However, solving these PDEs is a computationally demanding task and, more importantly, it is not clear how to `reverse' them to generate images with  given properties. Hence, we take an alternative approach of  approximating a PDE solver using a neural network~\cite{Korneev1,Korneev2}. To train such an approximation, we build a training set of pairs $(I_i,d_i)$, $i=1,\ldots,n$,
where $I_i$ is an input of the network and $d_i$, obtained by solving the PDE given $I$, is its label. In this work, we use a special class of deep neural networks
--- binarized neural networks ($\BNN$) that admit an exact encoding into a logical formula. We assume that $\Map$ is represented as a $\BNN$ and is given as part of input. We will elaborate on the training procedure  in Section~\ref{sec:exps}.

\section{The generative neural network approach}

One approach to tackle the constrained image generation problem is to use generative adversarial networks (GANs)~\cite{GoodfellowPMXWOCB14,RadfordMC15}. GANs are successfully used to produce samples of realistic images for commonly used datasets, e.g. interior design, clothes, animals, etc. A GAN can be described as a game between the image generator that produces synthetic (fake) images and a discriminator that distinguishes between fake and real images. The cost function is defined in such a way that the generator and the discriminator aim to maximize and minimize this cost function, respectively, turning the learning process into a minimax game between these two players. Each payer is usually represented as a neural network.  To apply GANs to our problem, we take a set of images $\{I_1,\ldots,I_n\}$ and pass them to the GAN. These images are samples of real images for the GAN. After the training procedure is completed, the generator network produces artificial images that look like real images. The main advantage of GANs is that it is a generic approach that can be applied to any type of images and can handle complex concepts, like animals, scenes, etc.\footnote{GANs exhibit well-known issues with poor convergence that we did not observe as our dataset is quite simple~\cite{Chintala}.} However, the main issue with this approach is that there is no way to explicitly pass declarative constraints into the training procedure.  One might expect that GANs are able to learn these constraints from the set of examples. However, this is not the case at the moment, e.g., GANs cannot capture counting constraints, like four legs, two eyes, etc.~\cite{Goodfellow17}.
Figure~\ref{fig:images} shows examples of images that GAN produces on a dataset with two grains per image. As can be seen from these examples, GAN produces images with an arbitrary number of grains between 1 and 5 per image. In some simple cases, it is easy to filter
wrong images. If we have more sophisticated constraints like convexity or size of grains, then most images will be invalid.
On top of this, to take into account process constraints, we need additional restrictions on the training procedure.  Overall, it is an interesting research question how to extend the GAN training procedure with physical constraints, which is beyond the scope of this paper~\cite{Oliveira}.
Next we consider our approach to the image generation problem.

\section{The constraint-based approach}
The main idea behind our approach is to encode the image generation problem as a logical formula.
To do so, we need to encode all problem constraints and the mapping between an image and its label
as a set of constraints. We start with constraints that encode an approximate PDE solver.
We denote $[N]$ a range of numbers from $1$ to $N$.

\subsection{Approximation of a PDE solver.}
One way to approximate a diffusion PDE solver is to use a neural network~\cite{Korneev1,Korneev2}.
A neural network is trained on a set of binary images $I_i$  and their labels $d_i$, $i=1,\ldots,n$.
During the training procedure, the networks
takes an image $I_i$  as an input and outputs its estimate of the parameter vector $\hat{d}_i$.
As we have ground truth parameters $d_i$ for each image, we can use the mean square error or absolute value error as a cost function to perform optimization~\cite{Narodytska}.
In this work, we take the same approach. However, we use a special
type of networks: Binarized Neural Networks (\BNN). $\BNN$ is a feedforward network where weights and activations are  binary~\cite{BNNNIPS2016}. It was shown in~\cite{Narodytska,Cheng2017} that $\BNN$s allow exact encoding as logical formulas, namely, they can be encoded a set of reified linear constraints over binary variables.
We use $\BNN$s as they have a relatively simple structure and decision procedures scale to reason about small and medium size networks of this type.
In theory, we can use any exact encoding to represent a more general network, e.g., MILP encodings that are used to check robustness properties of neural networks~\cite{Katz2017,Cheng2017a}.
However, the scalability of decision procedures is the main limitation in the use of more general networks.
We use the ILP encoding as in~\cite{Narodytska} with a minor modification of the last layer as we have numeric outputs instead of categorical outputs. We denote $\BBNN(I,d)$ a logical formula that encodes $\BNN$ using reified linear constraints over Boolean variables (Section 4, ILP encoding~\cite{Narodytska}).

\subsection{Geometric and process constraints.}
\begin{figure}
\centering
\includegraphics[width=1\linewidth]{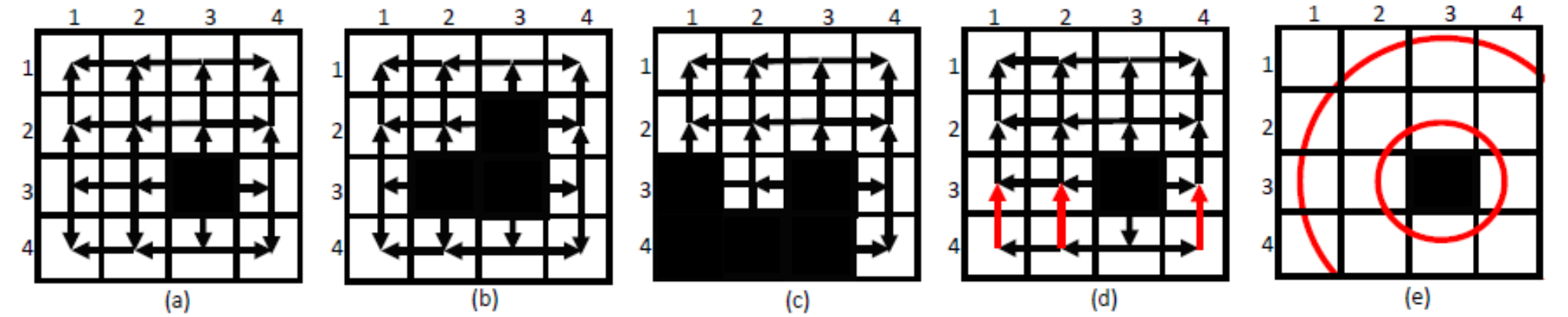}
\caption{Illustrative examples of additional structures used by constraint-based model.}
\label{fig:examples}
\end{figure}

Geometric constraints can be roughly divided into three types.
The first type of constraints defines the high-level structure
of the image. The high-level structure of our images is defined by the number of grains
present in the image.  Let $w$ be the number of grains per image.
We define a grid of size $t\times t$. Figure~\ref{fig:examples}(a) shows
an example of a grid of size $4 \times 4$. We refer to a cell $(i,j)$ on the grid as a pixel
as this grid encodes an image of size $t\times t$.
Next we define the neighbor relation on the grid. We say that a cell $(h,g)$
is a neighbour of $(i,j)$ if these cells share a side. For example, $(2,3)$
is a neighbour of $(2,4)$ as the right side of  $(2,3)$ is shared with $(2,4)$.
Let $\neub(i,j)$ be the set of neighbors of $(i,j)$ on the gird.
For example, $\neub(2,3) = \{(1,3),(2,2),(2,4), (3,3)\}$.

\vspace{-20pt}
\paragraph{Variables.} For each cell we introduce a Boolean variable $c_{i,j,r}$, $i,j \in [t]$, $r \in [w+1]$.
$c_{i,j,r} = 1$ iff the cell $(i,j)$ belongs to the $r$th grain, $r =1,\ldots, w$.
Similarly, $c_{i,j,w+1} = 1$ iff  the cell $(i,j)$ represents a void area.
\paragraph{Each cell is either a black or white pixel.} We enforce that each cell contains  either a grain or a void area.
\begin{equation}\label{eq:onegrain}
  \begin{array}{lcr}
    \sum_{r=1}^{w+1} c_{i,j,r}  = 1 & \quad\quad & j, i \in [t]\\
  \end{array}
\end{equation}

\paragraph{Grains do not overlap.}
Two cells that belong to different grains cannot be neighbours.
\begin{equation}\label{eq:overlap}
  \begin{array}{lcr}
    c_{i,j,r} \rightarrow \neg c_{h,g,r'} & \quad\quad & (h,g) \in \neub(i,j), r' \in [w]\setminus\{r\}
  \end{array}
\end{equation}

\paragraph{Grains are connected areas.} We enforce connectivity constraints for each grain.
By connectivity we mean that there is a path between two cells of the same grain using only cells that belong
to this grain. Unfortunately, enforcing connectivity constraints is
very expensive. Encoding the path constraint results in a prohibitively large encoding.
To deal with this explosion, we restrict the space of possible grain shapes.
First, we assume that we know the position of one pixel of this grain that we pick randomly.
Let $s_r = (i,j)$ be a random cell, $r \in [w]$.
Then we implicitly build a directed acyclic graph (DAG) $G$ starting from this cell $s_r$ that covers the entire grid.
Each cell of a grid is a node in this graph. The node that corresponds to the cell $s_r$ does not have incoming arcs.
There are multiple ways to build a $G$ from $s_r$.  Figure~\ref{fig:examples}(a) and (d) show two possible ways to build a
DAG that covers a grid starting from cell $(3,3)$. 
Next we define a parent relation in $G$. Let $\prt_G(i,j)$ be the set of parents of cell $(i,j)$ in $G$. For example,
$\prt_G(2,2) = \{(2,3), (3,2)\}$ in our example on  Figure~\ref{fig:examples}(a). Given a DAG $G$, we can easily enforce connectivity relation w.r.t. $G$.
The following constraint ensures that a cell $(i,j)$ belongs to the $r$th grain iff one of its parents in $G$
belongs to the same grain. Moreover, by enforcing  connectivity constraints on the void area, we make sure that grains do not contain isolated void areas inside them.
\begin{equation}\label{eq:connectedgrain}
  \begin{array}{lll}
    (c_{i,j,r}),  & \quad\quad & s_r = (i,j), r \in [w+1],\\
    \left(\wedge_{(h,g)\in \prt_G(i,j)}\neg c_{h,g,r}\right)\rightarrow \neg c_{i,j,r},   & \quad\quad &  j, i \in [t], r \in [w+1]\\
  \end{array}
\end{equation}

Given a DAG $G$, we can generate grains of multiple shapes. For example, Figure~\ref{fig:examples}(b) shows one possible grain. However, we also lose some valid shapes that are ruled out by the choice of graph $G$. For example, Figure~\ref{fig:examples}(c) gives an example of a shape that is not possible to build using $G$ in Figure~\ref{fig:examples}(a).
However, if we select a different random DAG $G'$, e.g., Figure~\ref{fig:examples}(d), then this shape is one of the possible shapes
for $G'$. In general, we can pick $s_r$ and DAG randomly, it is possible to generate a variety of shapes.

\paragraph{Compactness of a grain.} The second set of constraints is about restrictions on a single grain. The compactness constraint is a form of convexity constraint. We want to ensure that any two boundary points of a grain are close to each other. The reason for this constraint is that grains are unlikely to have a long snake-like appearance as solid particles tend to group together.
Sometimes, we need to enforce the convexity constraint, which is an extreme case of compactness. To enforce this constraint,
we again trade-off the variety of shapes and the size of the encoding. Now we assume that $s_r$ is the center of the grain.
Then we build virtual circles around this center that cover the grid. Figure~\ref{fig:examples}(e) shows examples of such circles.
Let $C_r(i,j) = \{C_r^1,\ldots, C_r^q\}$ be a set of circles that are built with the cell $s_r$ as a center.
The following constraint enforces that a cell that belongs to the circle $C_r^v$ can be in the $r$th grain iff all cells
from the inner circle $C_r^{v-s}$ belong to the $r$th grain, where  $s$ is a parameter.
\begin{equation}\label{eq:convexgrain}
  \begin{array}{lcr}
    \vee_{c_{h,g,r} \in C_r^{v-s}} \neg c_{h,g,r} \rightarrow \neg c_{i,j,r}   & \quad\quad & c_{i,j,r} \in C_r^v, v \in [q],  r \in [w]\\
  \end{array}
\end{equation}
Note that if $s=1$ then we generate convex grains. In this case, every pixel from $C_r^v$ has to belong to the $r$th grain before we can add a pixel from the circle $C_r^{v+1}$ to this grain.
\paragraph{Boundary constraints.}
We also have a technical constraint that all cells on the boundary of the grid must be void pixels. They are required to define boundary conditions for PDEs on generated images.
\begin{equation}\label{eq:boundary}
  \begin{array}{lcr}
    (c_{i,j,w+1})   & \quad\quad & j = t \vee  i = t\\
  \end{array}
\end{equation}
\paragraph{Connecting with $\BNN$.}
We need to connect variables $c_{i,j,r}$ with the inputs of the network.
\begin{equation}\label{eq:boundary2}
  \begin{array}{lcr}
    c_{i,j,r} \rightarrow I_{i,j} =1 & \quad\quad & j, i \in [t],  r \in [w],\\
    c_{i,j,w+1} \rightarrow I_{i,j} =0 & \quad\quad & j, i \in [t]. \\
  \end{array}
\end{equation}

\paragraph{Process constraints.}
Process constraints are enforced on the output of the network.
Given ranges $[a_i,b_i]$, $i\in [m]$ we have:
\begin{equation}\label{eq:process}
  \begin{array}{lcr}
    a_i \leq  d_{i}  \leq b_i & \quad\quad & i \in [m]
  \end{array}
\end{equation}

\paragraph{Summary.}
To solve the constrained random image generation problem, we solve the conjunctions of
constraints~\eqref{eq:onegrain}--\eqref{eq:process} together with our ILP encoding $\BBNN(I,d)$. Randomness comes from the random seed
that is passed to the solver, a random choice of $s_r$ and $G$.

\section{Experiments}\label{sec:exps}
We conduct a set of experiments with our constraint based approach. We ran our experiments on  Intel(R) Xeon(R) 3.30GHz.  We use the timeout of 600 sec in all runs.

\paragraph{Training procedure.} We use two datasets, $D_2$ with 10K images  and $D_3$ with 5K images.
Each image in $D_2$ contains two grains and each image in $D_3$ contains three  grains. These images were labeled with dispersion
coefficients along the $x$-axis which is a number between 0.4 and 1. We performed  quantization on the dispersion coefficient value  to
map $d$ into an interval of integers between $40$ and $100$.
We use  mean absolute error ($MAE$) to train $\BNN$. $\BNN$ consists of three blocks with 100 neurons per layers
and one output. The $MAE$ is 4.2 for $D_2$ and 5.1  for $D_3$. We lose  accuracy compared to non-binarized networks,
e.g,  $MAE$  for the same non-binarized network is 2.5 for $D_2$. However, $\BNN$s are much easier to reason about, so we work
with this subclass of networks.

\paragraph{Image generation.}We use CPLEX and the SMT solver Z3 to solve instances produced by constraints~\eqref{eq:onegrain}--\eqref{eq:process} together with $\BBNN(I,d)$. In principle, other solvers could be evaluated on these instances. The best mode for Z3 was to use an SMT core based on CDCL and a theory solver for \emph{nested} Pseudo-Boolean and cardinality constraints. We noted that bit-blasting into sorting circuits did not scale, and Z3’s theory of linear integer arithmetic was also inadequate.
We considered six process constraints for $d$, namely, $d \in [a,b]$, $ [a,b] \in \{[40,50),\ldots, [90,100]\}$.
For each interval $[a,b]$,  we generate 100 random constrained problems.
The randomization comes from a random seed that is passed to the solver, the position of centers of each grain and the  parameter $s$ in the constraint~\eqref{eq:convexgrain}.
We used the same DAG $G$ construction as in Figure~\ref{fig:examples}(a) in all problems. Table~\ref{table:solved} shows summary of our results for CPLEX and Z3 solvers.
As can be seen from this table, these instances are relatively easy for the CPLEX solver.
It can solve most of them within the given timeout. 
The average time for $D_2$ is  25s and for $D_3$ is 12s with CPLEX.
Z3 handles most benchmarks, but we observed it gets stuck on examples that are very easy for CPLEX, e.g. the interval $[80,90)$ for $D_2$. We hypothesize that this is due to how watch literals are tracked in a very general way on nested cardinality constraints (Z3 maintains a predicate for each nested PB constraint and refreshes the watch list whenever the predicate changes assignment), when one could instead exploit the limited way that CPLEX allows conditional constraints.
The average time for $D_2$ is 94s and for $D_3$ is 64s with Z3.
\begin{wrapfigure}{r}{0.35\textwidth}
 \vspace{-20pt}
  \begin{center}
    \includegraphics[width=0.35\textwidth]{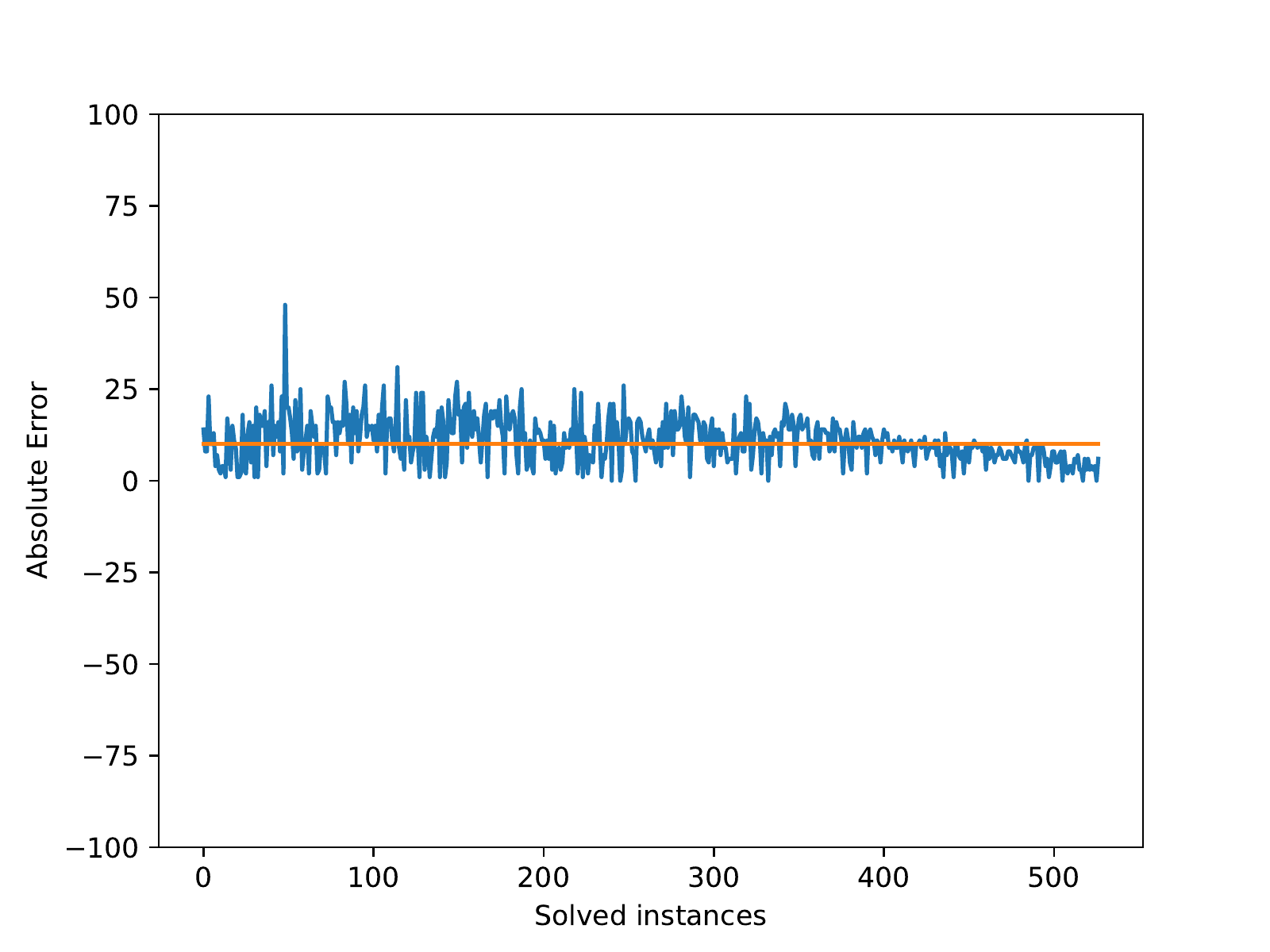}
  \end{center}
  \caption{The absolute error between $d$ and its true value. \label{fig:mae}}
  \vspace{-20pt}
\end{wrapfigure}
Figures~\ref{fig:images}(c)--(e) show examples of generated images for ranges $[40,50)$, $[60,70)$ and $ [90,100]$
for $D_2$ (the top row) and $D_3$ (the bottom row). For the process we consider, as the value of the dispersion coefficient  grows,
the black area should decrease as there should be fewer grain obstacles for a flow to go through the porous medium.
Indeed, images in Figures~\ref{fig:images}(c)--(e) follow this pattern, i.e. the black area on images with $d \in [40,50)$ is significantly larger
than on images with $d \in [90,100]$. Moreover, by construction, they satisfy geometric constraints that GANs cannot handle.
For each image we generated, we run a PDE solver to compute the true value of the dispersion coefficient on this image.
Then we compute the absolute error between the value of $d$ that our model computes and the true value of the coefficient.
Figure~\ref{fig:mae} shows absolute errors for all benchmarks that were solved by CPLEX.
First, this figure shows that our model
generates images with given properties.
The mean absolute error is about 10 on these instances.
Taking into account  that $\BNN$ has $MAE$ of  4.2 on $D_2$,  $MAE$ of 10 on new generated instances is a reasonable result.
Ideally, we would like $MAE$ to be zero. However, this error depends purely on the $\BNN$ we used.
To reduce this error, we need to improve the accuracy of $\BNN$ as
it serves as an approximator of a PDE solver. For example, we can use more binarized layers or use additional non-binarized layers.
Of course, increasing the power of the network leads to computational challenges solving the corresponding logical formulas.
%
%
%
%
\vspace{-10pt}
\begin{table}
\scriptsize
  \centering
  \begin{tabular}{|c|c|c|c|c|c|c||c|c|c|c|c|c|}
    \hline
   \multirow{2}{*}{Solver}  & \multicolumn{6}{c||}{$D_2$} & \multicolumn{6}{c|}{$D_3$} \\
    \cline{2-13}
    & [40,50)& [50,60)& [60,70)& [70,80) & [80,90)& [90,100]& [40,50)& [50,60)& [60,70)& [70,80) & [80,90)& [90,100]\\
    \hline
    CPLEX & 100 &99 & 99 & 98 & 100  & 41 &  100& 100 & 96 &99 & 100 & 84 \\
    Z3 &  98 &  89 & 81  &   74 &  56 &  12 & 100 & 97 & 97 &  97 &  96 &  54\\

    \hline
  \end{tabular}
  \caption{The number of solved instances in each interval $[a,b]$.}\label{table:solved}
\end{table}
\vspace{-35pt}
\section{Related work}
There are two lines of work related to our paper. The first one uses constraint
to enhance machine learning techniques with declarative constraints,
e.g. in solving constrained clustering problems and in data mining techniques that handle domain specific  constraints~\cite{DaoDV17,GanjiBS17,GunsDNTR17}.
One recent example is the work of  Ganji \emph{et al.}~\cite{GanjiBS17} who proposed a logical model for constrained community detection.
The second line of research explores embedding of domain-specific constraints in the GAN training procedure~\cite{Oliveira,Oliveira1,HuGLXBVN17,Osokin,RavanbakhshLMSP17}.
Work in this area is targeting various applications in physics and medicine that impose constraints, like
sparsity constraints, high dynamic range requirements (e.g. when pixel intensity in an image varies by orders of magnitude),
location specificity constraints (e.g. shifting pixel locations can change important image properties), etc.
However, this research area is emerging and the results are still preliminary.

\section{Conclusion}
In this paper we considered the constrained image generation problem for a physical process.
We showed that this problem can be encoded as a logical formula over Boolean variables.
For small porous media, we show that the generation process is computationally feasible for
modern decision procedures.
There are a lot of interesting future research directions.
First, the main limitation of our approach is scalability, as we
cannot use large networks with a number of weights in the order of
hundreds of thousands, as it is required by industrial applications.
However, constraints that are used to encode, for example,
binarized neural networks are mostly pseudo-Boolean constraints with
unary coefficients. Hence, it would be interesting to design
specialized procedures to deal with this fragment of constraints.
Second, we need to investigate different types of neural networks that
admit encoding into SMT or ILP. For instance, there is a lot of work
on quantized networks that use a small number of bits to encode each
weight, e.g.~\cite{DengJPWL17}. Finally, can we use similar techniques to reveal
vulnerabilities in neural networks?  For example, we might be able to
generate constrained adversarial examples or other special types of
images that expose undesired network behaviour.

\bibliography{lit}
\bibliographystyle{splncs}
\end{document}